%% file: main.tex
\documentclass[11pt]{article}

\usepackage[preprint]{acl}
\usepackage{times}
\usepackage{latexsym}
\usepackage[T1]{fontenc}
\usepackage[utf8]{inputenc}
\usepackage{microtype}
\usepackage{inconsolata}
\usepackage{graphicx}

\usepackage{hyperref}
\hypersetup{
    colorlinks,
    linkcolor={blue!80!black},
    citecolor={blue!80!black},
    urlcolor={blue!80!black}
}
\usepackage{url}
\usepackage{amssymb}
\usepackage{pifont} 
\usepackage{subcaption}
\usepackage{xspace}

\usepackage{enumitem}
\usepackage{algorithm}
\usepackage{xcolor}
\usepackage{algpseudocode}
\usepackage[normalem]{ulem}
\usepackage[most]{tcolorbox}

\usepackage{amsmath, amsfonts}
\usepackage{nicefrac}
\PassOptionsToPackage{table}{xcolor}
\usepackage{booktabs}
\usepackage{colortbl}  
\usepackage{multirow}
\usepackage{adjustbox}
\usepackage{hyperref}
\usepackage{tabularx}
\usepackage{hhline}
\usepackage{makecell}
\usepackage{arydshln}
\usepackage{array}
\usepackage{colortbl}
\usepackage{wrapfig}
\usepackage{caption}
\usepackage{booktabs}       
\usepackage{amsfonts}       
\usepackage{nicefrac}       
\usepackage{setspace}

\usepackage{placeins}
\definecolor{light_origin}{RGB}{207, 117, 58}
\definecolor{cell_purple}{RGB}{76, 141, 187}
\definecolor{small_blue}{RGB}{27, 63, 125}
\definecolor{small_origin}{RGB}{94, 41, 16}
\definecolor{mypink}{HTML}{D92879}

\definecolor{light_purple}{RGB}{235,236,242}

\definecolor{light_blue}{RGB}{244,249,254}

\newcounter{tcbtable}

\newcolumntype{B}{>{\columncolor{blue!4}}c}
\newcolumntype{d}{>{\columncolor{brown!4}}c}
\newcolumntype{q}{>{\columncolor{green!2}}c}
\newcolumntype{R}{>{\columncolor{red!4}}l}     
\newcolumntype{P}{>{\columncolor{purple!4}}c}   
\newcolumntype{Y}{>{\columncolor{yellow!4}}c}   

\definecolor{bgcolor}{RGB}{242, 243, 245} 
\makeatletter
\def\adl@drawiv#1#2#3{%
        \hskip.5\tabcolsep
        \xleaders#3{#2.5\@tempdimb #1{1}#2.5\@tempdimb}%
                #2\z@ plus1fil minus1fil\relax
        \hskip.5\tabcolsep}
\newcommand{\cdashlinelr}[1]{%
  \noalign{\vskip\aboverulesep
            \global\let\@dashdrawstore\adl@draw
            \global\let\adl@draw\adl@drawiv}
  \cdashline{#1}
  \noalign{\global\let\adl@draw\@dashdrawstore
            \vskip\belowrulesep}}

\tcbset{
  aibox/.style={
    width=\linewidth,
    top=8pt,
    bottom=4pt,
    colback=blue!6!white,
    colframe=black,
    colbacktitle=black,
    enhanced,
    center,
    attach boxed title to top left={yshift=-0.1in,xshift=0.15in},
    boxed title style={boxrule=0pt,colframe=white,},
  }
}
\newtcolorbox{AIbox}[2][]{aibox,title=#2,#1}
\definecolor{lightblue}{rgb}{0.22,0.45,0.70}%

\PassOptionsToPackage{table}{xcolor}
\usepackage{booktabs}
\usepackage{placeins}
\definecolor{light_origin}{RGB}{207, 117, 58}
\definecolor{cell_purple}{RGB}{76, 141, 187}
\definecolor{small_blue}{RGB}{27, 63, 125}
\definecolor{small_origin}{RGB}{94, 41, 16}
\definecolor{mypink}{HTML}{D92879}

\newcommand{\method}{\textsc{SCR}\xspace}

\tcbset{
    training_data/.style={
        colback=blue!5,      
        colframe=black,      
        sharp corners,       
        left=3pt, right=3pt, top=3pt, bottom=3pt,
        boxrule=1pt,
        fontupper=\normalsize,
        parbox=false,        
        enhanced,
        borderline={0.5pt}{0pt}{black}
    }
}

\title{Structured Reasoning for Large Language Models}
\author{
  Jinyi Han\textsuperscript{\rm $\heartsuit$}\footnotemark[1],
  Zixiang Di\textsuperscript{\rm $\heartsuit$}\thanks{Equal contribution. Work done during an internship at ByteDance.},
  Zishang Jiang\textsuperscript{\rm $\clubsuit$},
  \\
{\bfseries
  Ying Liao\textsuperscript{\rm $\clubsuit$},
  Jiaqing Liang\textsuperscript{\rm $\clubsuit$},
  Yongqi Wang\textsuperscript{\rm $\spadesuit$},
  Yanghua Xiao\textsuperscript{\rm $\clubsuit$}\thanks{Corresponding author.}
  }
  \\
  \textsuperscript{\rm $\heartsuit$} East China Normal University,\;
  \textsuperscript{\rm $\clubsuit$} Fudan University,\;
  \textsuperscript{\rm $\spadesuit$} Beijing Institute of Technology
}
\begin{document}
\maketitle
\begin{abstract}
Large language models (LLMs) achieve strong performance by generating long chains of thought, but longer traces always introduce redundant or ineffective reasoning steps. One typical behavior is that they often perform unnecessary verification and revisions even if they have reached the correct answers. This limitation stems from the unstructured nature of reasoning trajectories and the lack of targeted supervision for critical reasoning abilities. To address this, we propose \textbf{\underline{S}}tru\textbf{\underline{C}}tured \textbf{\underline{R}}easoning (\method), a framework that decouples reasoning trajectories into explicit, evaluable, and trainable components. We mainly implement \method using a \textit{Generate–Verify–Revise} paradigm. Specifically, we construct structured training data and apply Dynamic Termination Supervision to guide the model in deciding when to terminate reasoning. To avoid interference between learning signals for different reasoning abilities, we adopt a progressive two-stage reinforcement learning strategy: the first stage targets initial generation and self-verification, and the second stage focuses on revision. Extensive experiments on three backbone models show that \method substantially improves reasoning efficiency and self-verification. Besides, compared with existing reasoning paradigms, it reduces output token length by up to 50\%.
\end{abstract}

\section{Introduction}
\vspace{-0.2cm}
\input{subsection/14_introduction}

\section{Pilot Experiments}
\vspace{-0.2cm}
\input{subsection/22_pilot_experiments}

\section{Methods}
\vspace{-0.2cm}
\input{subsection/35_methods}

\section{Experiments}
\vspace{-0.2cm}
\input{subsection/42_exp}

\vspace{-0.2cm}
\section{Related Work}
\vspace{-0.2cm}
\input{subsection/51_related_work}
\vspace{-0.2cm}
\section{Conclusion}
\vspace{-0.2cm}
In this paper, we propose StruCtured Reasoning (\method), a framework that organizes reasoning into an explicit, assessable, and trainable form. We implement it follows the Generate–Verify–Revise paradigm. First, we apply DTS to enable the model to produce structured reasoning traces and determine reasoning termination based on self-verification. We then further enhance critical reasoning abilities through a two-stage reinforcement learning strategy. Extensive experiments on several benchmarks demonstrate that \method consistently improves reasoning performance across tasks.
\section*{Limitations}
While the structured reasoning framework naturally supports multi-round verification and revision, our experiments are restricted to two correction rounds. This setting serves as a minimal yet sufficient configuration to evaluate the core behaviors of critique and revision. Extending to more rounds requires synthesizing high-quality multi-round correction trajectories, whose cost increases rapidly with the number of rounds. We leave the exploration of more cost-efficient strategies for multi-round data construction to future work.

\bibliography{main}

\clearpage
\appendix
\section*{Appendix}
\label{sec:appendix}
\input{subsection/appendix_v1}

\end{document}

%% file: subsection/14_introduction.tex
Large language models (LLMs) achieve strong performance on complex reasoning tasks by generating long chains of thought~\citep{min2024imitate}. Models such as OpenAI o1~\citep{jaech2024openai}, DeepSeek-R1~\citep{guo2025deepseek}, and the Qwen Thinking series~\citep{yang2025qwen3} adopt a System-2~\citep{frankish2010dual} like reasoning style, producing answers through deliberate multi-step inference. This explicit reasoning paradigm improves decision quality and substantially enhances performance on mathematical reasoning~\citep{shao2024deepseekmath} and code generation~\citep{jiang2024survey}.
\begin{figure}[t]
    \centering
\includegraphics[width=.99\linewidth]{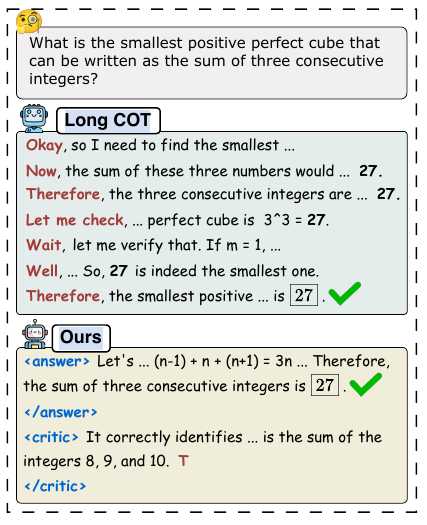}
    \vspace{-0.5cm}
 \caption{\textbf{Comparison of reasoning trajectories.}
\textit{Long CoT:} A single reasoning process in which all thinking operators are blended together, resulting in redundant self-verification loops even after a correct solution is reached.
\textit{Ours:} A structured reasoning trajectory with explicitly separated components, where each stage is verifiable and trainable. Generation stops once correctness is confirmed.}
\label{fig:intro}
\end{figure}
\begin{figure*}
    \centering
\includegraphics[width=0.99\linewidth]{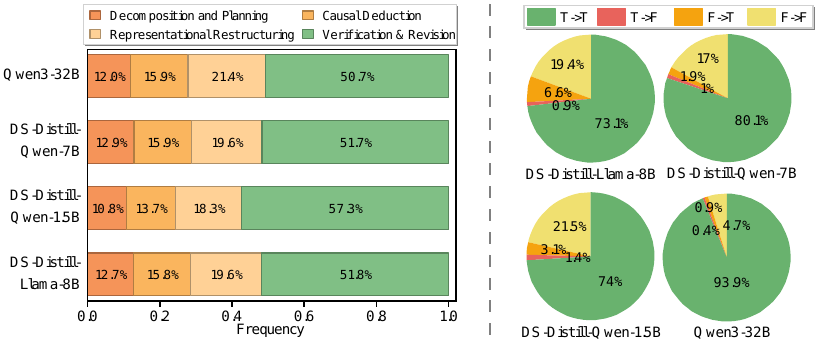}
    \vspace{-0.2cm}
    \caption{\textbf{Behavioral analysis of long-chain reasoning in large reasoning models.}
\textit{Left:} Frequency distribution of different thinking operators within long CoT trajectories.
\textit{Right:} Effectiveness of revision behaviors in long CoT, quantified by the transition in answer correctness pre- and post-revision.}
    \vspace{-0.2cm}
    \label{fig:pilot}
\end{figure*}

Despite these gains, longer reasoning traces do not necessarily guarantee high-quality reasoning~\citep{chen2024not,luo2025o1}, largely due to the model’s limited self-verification ability~\citep{zhang2025understanding}. Even after producing a correct solution, it may repeatedly verify its reasoning or make unnecessary revisions~\citep{kang2025first,zeng2025revisiting}, as illustrated in Figure~\ref{fig:intro}. Most importantly, These verification and revision behaviors dominate the reasoning trajectory (Figure~\ref{fig:pilot} Left) but rarely yield meaningful corrections (Figure~\ref{fig:pilot} Right). When such high-frequency behaviors fail to produce gains, they degenerate into cognitive idling, where the model continues reasoning without substantive progress.

This issue can be traced to the unstructured nature of long reasoning trajectories under commonly used outcome-based optimization. In existing large reasoning models, different reasoning operations are intertwined within a single trajectory without clear structured boundaries~\citep{team2025kimi}. Furthermore, prevailing training paradigms, such as Reinforcement Learning with Verifiable Rewards (RLVR)~\citep{kirk2024understanding, chu2025sftmemorizesrlgeneralizes, betley2025emergent}, optimize exclusively for final answer correctness. Such sparse, global reward signals provide little guidance for assessing intermediate steps, including whether verification is warranted or whether a revision effectively rectifies an error. As a result, these critical atomic abilities cannot be independently distinguished, evaluated, or trained.

Therefore, it is essential to restructure the reasoning trajectory into a structured form, where each critical reasoning ability is associated with controllable and assessable learning signals. To this end, we propose \textbf{\underline{S}}tru\textbf{\underline{C}}tured \underline{\textbf{R}}easoning (\method), a framework that allows each critical ability to be independently evaluated and trained. In our implementation, we achieve this by decoupling the reasoning trajectory into the key abilities of initial generation, verification, and revision, and applying targeted optimization to each ability.

Specifically, we first construct structured reasoning training data during supervised fine-tuning (SFT) to guide the model toward a \textit{Generate–Verify–Revise} paradigm. We then introduce Dynamic Termination Supervision, which conditions termination decisions on the outcomes of self-verification, ensuring that verification and revision are applied appropriately. Building on this foundation, we employ a two-stage reinforcement learning strategy, optimizing each reasoning ability independently while preventing interference between their learning signals.
The first stage focuses on improving initial generation and self-verification. The second stage focuses on revision and applies fine-grained learning signals to correct errors more effectively.

In summary, our contributions are three-fold:
\begin{itemize}[wide=1pt, nosep]
    \item We reformulate long-chain reasoning as a structured reasoning trajectory, a general framework for organizing reasoning steps. In our implementation, we explicitly decouple generation, verification, and revision, enabling each critical reasoning ability to be independently evaluated and trained.
    \item We introduce \emph{Dynamic Termination Supervision} in SFT, enabling the model to autonomously decide when to stop reasoning based on self-verification outcomes.
    \item We propose a two-stage reinforcement learning framework with fine-grained rewards to target the optimization of each reasoning ability.
\end{itemize}

%% file: subsection/22_pilot_experiments.tex
To examine how large reasoning models (LRM) allocate their computational budget during long-chain reasoning, we conduct a pilot study based on \textit{Information-Processing Theory}~\citep{simon1978information}. This theory views problem solving as a search process driven by a set of thinking operators. Following this view, we categorize the reasoning behaviors into four types and analyze how these operators are used by LRMs. Experimental details are provided in Appendix~\ref{sec:pilot}.

We observe that verification and revision overwhelmingly dominate the reasoning process across models. As shown in Figure~\ref{fig:pilot} Left, their execution frequency substantially exceeds that of decomposition, causal deduction, and representational restructuring across four LRMs. This imbalance indicates that a large portion of the computational budget in long reasoning is allocated to post-hoc verification. 

However, a closer analysis of revision outcomes reveals a clear mismatch between frequency and effectiveness. As shown in Figure~\ref{fig:pilot} Right, the majority of revision steps preserve the correctness state of the initial answer. Over 90\% of revisions either confirm already correct answers or leave incorrect answers unchanged (T$\rightarrow$T or F$\rightarrow$F), while successful corrections from incorrect to correct account for only a small fraction of cases (F$\rightarrow$T). This pattern indicates that revision is rarely triggered by accurate error detection, but instead functions as repetitive confirmation.

\vspace{-0.2cm}
\begin{AIbox}{Highlights}
Overall, these results suggest that although verification and revision dominate the reasoning trajectory, they achieve a low error-correction success rate.
\end{AIbox}
\vspace{-0.3cm}
\begin{figure*}[ht]
    \centering
    \includegraphics[width=0.99\linewidth]{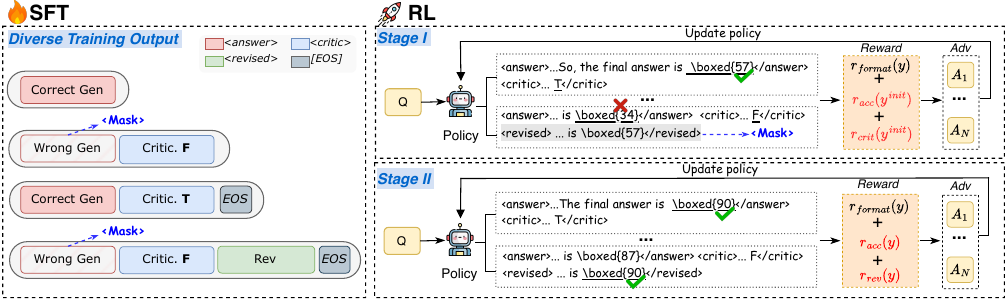}
    \vspace{-0.2cm}
\caption{\textbf{Overview of the proposed training framework.} 
\textit{Left:} SFT with Dynamic Termination Supervision and Selective Loss Masking.
\textit{Right:} Staged reinforcement learning with stage-specific reward designs for learning initial generation, self-verification, and revision.}
    \vspace{-0.2cm}
    \label{fig:method}
\end{figure*}

%% file: subsection/35_methods.tex
\subsection{Problem Formulation}
In this work, we formalize problem-solving as a single-pass generation of a structured reasoning trajectory. In this trajectory, initial generation, verification, and revision are represented as ordered components, allowing each capability to be learned, evaluated, and optimized during training.

Formally, given an input problem $x$, the reasoning process is represented as a trajectory:
\vspace{-0.2cm}
\begin{equation}
y = \left( y^{(0)}, c^{(0)}, y^{(1)}, \dots, c^{(T-1)}, y^{(T)} \right),
\end{equation}
\vspace{-0.1cm}
where $y^{(0)}$ denotes the initial response, and $y^{(T)}$ is the final answer. At each iteration $t \in \{1, \dots, T\}$, the model produces a critique $c^{(t)}$, which explicitly assesses the correctness of the answer $y^{(t)}$, and when incorrect, identifies the key errors
that need to be corrected. Based on this critique, the model either revises the answer to produce $y^{(t+1)}$ or keeps the previous answer if no correction is needed. We model this process autoregressively by factorizing the probability of a reasoning trajectory as:
\vspace{-0.2cm}
\begin{equation}
\begin{aligned}
p(y \mid x) = & p(y^{(0)} \mid x) \prod_{t=0}^{T-1} p(c^{(t)} \mid x, y^{(t)}) \cdot 
\\ &p(y^{(t+1)} \mid x, y^{(t)}, c^{(t)}).
\end{aligned}
\end{equation}
\vspace{-0.2cm}

\subsection{Behavior Cloning on Structured Reasoning Trajectories}
\subsubsection{Data Synthesis}
To construct high-quality supervision signals for structured reasoning, we develop an automated data synthesis pipeline that provides explicit training data for three critical stages. 

For each problem, we first sample multiple candidate solutions to cover diverse reasoning behaviors. These candidates are automatically labeled by comparing their final answers with the ground-truth solution, yielding both correct and incorrect initial responses.

Given each labeled response, teacher models generate a corresponding critique, which either validates a correct solution or identifies errors in an incorrect one. When errors are detected, the teacher further produces a targeted revision to correct the original response. All synthesized trajectories are retained only if the final answer is verified to be correct. Further details of the data construction process are provided in the Appendix~\ref{sec:data_construction}.

Based on the synthesis pipeline described above, the SFT training data consists of two types of structured reasoning forms. The first type, \textbf{Correct-Answer Trajectories} contain cases where the initial response is correct and subsequent critiques only verify its correctness, resulting in no modification to the answer. The second type, \textbf{Correction Trajectories} contain cases where the initial response is incorrect and is corrected through one or more critique–revision rounds.

To make the internal reasoning stages explicit, each component in the training data is enclosed within paired start and end tags: initial answers are placed between \textit{<answer>} and \textit{</answer>}, critique content between \textit{<critic>} and \textit{</critic>}, and revised outputs between \textit{<revised>} and \textit{</revised>}. An example of this data format is shown in Box ~\ref{tab:traing_data}.

\subsubsection{Supervised Fine-Tuning for Structured Reasoning}
We perform Supervised Fine-Tuning (SFT) to train the model to generate structured reasoning trajectories in a single pass.

\textbf{Dynamic Termination Supervision (DTS).} 
To enable adaptive generation control, we supervise the placement of the end-of-sequence (\textit{[EOS]}) token based on the critique outcome.
For samples where the critique confirms the initial answer is correct, an \textit{[EOS]} token is inserted immediately after the critique, signaling the model to stop generation.
For samples where errors are identified, no \textit{[EOS]} token is provided, forcing the model to continue generating a revision.
This design directly supervises the model to terminate only when correctness is verified and to proceed with correction otherwise. The training data are organized in the structured form illustrated in Figure~\ref{fig:method}.

\textbf{Selective Loss Masking (SLM).}
To prevent the model from learning incorrect reasoning patterns, we apply selective loss masking for trajectories with erroneous initial answers. Tokens in the incorrect initial answer are masked from the loss, while still serving as context for learning critique and revision.

Formally, the training objective is defined as:
\vspace{-0.2cm}
\begin{equation}
    \mathcal{L} = - \sum_{t=1}^{T} m_t \cdot \log P_\theta(y_t \mid x, y_{<t}),
\end{equation}
\vspace{-0.1cm}
where $m_t \in \{0, 1\}$ is a binary mask. For samples with incorrect initial answers, we set $m_t = 0$ for tokens in $y^{init}$ and $m_t = 1$ for tokens in $y^{c}$ and $y^{rev}$. This ensures that incorrect initial reasoning serves only as context for learning verification and correction, rather than as a supervision target.

\subsection{Staged Reinforcement Learning}
We apply reinforcement learning with Group Relative Policy Optimization (GRPO) to improve structured correction trajectories.
The trajectory comprises three abilities: initial generation, self-verification, and revision. Jointly optimizing all components is challenging due to entangled learning signals. Therefore, we adopt a two-stage training scheme. Stage I focuses on stabilizing initial reasoning and self-verification. Stage II then optimizes revision conditioned on reliable verification.

\subsubsection{Stage I: Enhancing Initial Generation and Self-Verification.}
Stage I aims to improve the model’s initial answer quality and its ability to accurately assess correctness.
Following the GRPO framework, for each input $x$, we sample a group of outputs and update the policy using group-relative advantages.

\textbf{Reward Design.}
Stage I adopts a rule-based reward scheme composed of three components: a format reward \(r_{format}\), an initial accuracy reward \(r_{acc}\), and a self-verification reward \(r_{crit}\).  
These rewards supervise output structure, correctness of the initial answer, and consistency between self-verification and actual correctness, respectively.

\textbf{\textit{Format Reward.}} It enforces a predefined output structure. Let \(C_i(y) \in \{0,1\}\) indicate whether output \(y\) satisfies constraint \(C_i\). We define five constraints:\\
\textbf{ \textbf{\(C_1\)}}: Output must include both \(\textit{<answer>}\) and \(\textit{<critic>}\) tag pairs; \\\textbf{\textit{\(C_2\)}}: \(\textit{<critic>}\) section must conclude with either \textit{T} or \textit{F}; \\\textbf{\textit{\(C_3\)}}: If the critique concludes with \textit{F}, a \(\textit{<revised>}\) tag must follow; \\\textbf{\textit{\(C_4\)}}: Final answers in \textit{<answer>} and \textit{<revised>} are enclosed in \textit{\textbackslash \textbackslash boxed\{\}};\\
\textbf{\textit{\(C_5\)}}: All tags preserve their relative order. \\
The format reward is defined as:
\begin{equation}
    r_{format}(y) = \prod_{i=1}^{5} C_i(y),
\end{equation}
which assigns a nonzero reward only to fully well-formed outputs.

\textbf{\textit{Initial Accuracy Reward.}} This reward measures whether the model’s initial answer \(y^{init}\) matches the ground truth \(y^\star\):
\begin{equation}
r_{acc}(y^{init}) =
\begin{cases}
1, & \text{if } y^{init} = y^\star, \\
0, & \text{otherwise}.
\end{cases}
\end{equation}

\textit{\textbf{Self-Verification Reward.}} This reward evaluates whether the model’s self-verification \(crit(y) \in \{T,F\}\) correctly reflects the actual correctness of its initial answer:
{\small
\begin{equation}
r_{crit}(y^{init}) =
\begin{cases}
1, & \text{if } (r_{\text{acc}}(y^{init})=1 \land crit(y^{init})=T) \\
   & \text{or } (r_{\text{acc}}(y^{init})=0 \land crit(y^{init})=F), \\
0, & \text{otherwise}.
\end{cases}
\end{equation}
}
\textbf{Total Reward.} The total reward in Stage I is a weighted sum of the three components:
{\small
\begin{equation}
r_{\text{total}}(y) = \alpha \cdot r_{\text{format}}(y) + \beta \cdot r_{acc}(y^{init}) + \gamma \cdot r_{crit}(y^{init}).
\end{equation}
}
where \(\alpha\), \(\beta\), and \(\gamma\) control the relative contributions.\\
\textbf{Optimization Objective.} 
During Stage I, gradient updates are applied only to tokens in the \textit{<answer>} and \textit{<critic>} segments, while the \textit{<revised>} block is excluded from optimization.  
This design isolates the learning of initial reasoning and self-verification from revision behaviors.

To implement this in GRPO, we modify the standard objective by masking the revision region. Let \(y^{\setminus \text{rev}}\) denote the tokens in the \textit{<answer>} and \textit{<critic>} segments, excluding the \textit{<revised>} block. 
Only tokens in \(y^{\setminus \text{rev}}\) are included in policy optimization.
The Stage I objective is then defined as:
\vspace{-0.2cm}
\begin{equation}
\begin{aligned}
J_{\text{Stage-I}}(\theta)=\mathbb{E}_{x \sim D},{a_i \sim \pi_\theta(x)} \bigg[ &\frac{1}{G} \sum_{i=1}^{G} A_i(y_i^{\setminus \text{rev}}|x) \cdot \\ \min \Big( r_i, \,\text{clip}(r_i, 1-\epsilon, 1+\epsilon)\Big) \bigg],
\end{aligned}
\label{eq:grpo}
\end{equation}
\vspace{-0.1cm}
where 
\(
r_i = \frac{\pi_\theta(y_i^{\setminus \text{rev}}|x)}{\pi_{\mathrm{old}}(y_i^{\setminus \text{rev}}|x)}
\), and \(A_i\) is the group-relative advantage computed from the total reward \(r_{\text{total}}(y_i)\).
\subsubsection{Stage II: Optimizing Revision}
After the model acquires stable initial reasoning and self-verification, Stage II optimizes the revision behavior.

\textbf{\textit{Revision-Focused Rewards.}} 
Stage II adopts a compound reward that combines a format reward \(r_{format}\), a final answer accuracy reward \(r_{acc}\), and a revision-specific reward \(r_{rev}\).
The format reward enforces structural validity as in Stage I, while the accuracy reward evaluates the correctness of the final output.

The revision reward directly measures the effectiveness of the correction step by comparing the correctness before and after revision. It is designed according to four principles: \\ 
\textit{\textbf{(i) }}reward successful correction of initially incorrect answers; \\
\textit{\textbf{(ii)}} assign mild penalties to ineffective revisions \\
\textit{\textbf{(iii)}} penalize unnecessary revisions when the initial answer is already correct; \\
\textit{\textbf{(iv)}} strongly penalize harmful revisions that degrade a correct answer.

Formally, the revision reward is defined as:
\begin{equation}
r_{rev}(y) =
\begin{cases}
\mu _1, & r_{acc}(y^{init}) = 0 \land r_{acc}(y) = 1 \\
\mu _2, & r_{acc}(y^{init}) = 0 \land r_{acc}(y) = 0 \\
\mu _3, & r_{acc}(y^{init}) = 1 \land r_{acc}(y) = 1 \\
\mu _4, & r_{acc}(y^{init}) = 1 \land r_{acc}(y) = 0 \\
0, & \text{no revision}
\end{cases}
\end{equation}
The coefficients satisfy the ordering \(\mu _1 > \mu _2 > \mu _3 > \mu _4\), reflecting the relative desirability of revision outcomes. 
\input{tables/main_result}

\textbf{Total Reward.} The total reward in Stage II is computed as:
\begin{equation}
r_{total}(y) = \nu \cdot r_{format}(y) + \eta \cdot r_{acc}(y) + \varphi  \cdot r_{rev}(y).
\end{equation}
where \(\nu\), \(\eta\), and \(\varphi\) weight the three rewards. 

\textbf{Optimization Objective.} 
In Stage II, we remove the masking strategy introduced in Stage I, enabling gradients to propagate through the entire reasoning trajectory. While retaining the same GRPO objective, the policy is now optimized over the full reasoning sequence.

%% file: tables/main_result.tex
\setlength\tabcolsep{5.7pt}
\begin{table*}[ht]
\centering
\def\arraystretch{.99}
\setlength{\tabcolsep}{0.42em}
\resizebox{1.0\linewidth}{!}{
\begin{tabular}{ll BBBBB ddd R}
\toprule
\multirow{2}{*}{\textbf{Models}} & \multirow{2}{*}{\textbf{Method}} 
& \multicolumn{5}{c}{\textbf{In-Domain}} & \multicolumn{3}{c}{\textbf{Out-of-Domain}} &\cellcolor{white}{\multirow{2}{*}{\textbf{AVG}}}\\

\cmidrule(lr){3-7} \cmidrule(lr){8-10} 
& & \cellcolor{white}{MATH500} & \cellcolor{white}{AIME24} & \cellcolor{white}{AIME25} & \cellcolor{white}{Olympiad} & \cellcolor{white}{AMC} &\cellcolor{white}{ARC}   & \cellcolor{white}{MMLU} & \cellcolor{white}{GPQA} \\ 
\midrule
\multirow{7}{*}{\textbf{Qwen2.5-7B-Instuct}} 
&Base&75.60 	&11.00 	&6.00 	&39.70 	&42.05 	&84.39 	&55.80 	&33.84	&43.55 \\
&Self-Refine &68.40 	&9.33 	&5.67 	&36.44 	&38.67 	&57.51 	&55.90 	&34.85	&38.35 \\
&GRPO &77.40 	&11.33 	&10.33 	&42.81 	&50.60 	&85.49 	&57.00 	&37.88	&46.60 \\
&SFT + GRPO &78.80 	&15.33 	&11.67 	&42.81 	&50.84 	&80.63 	&57.80 	&35.86	&46.72 \\
\cdashlinelr{2-11}
&\method-SFT &74.20 	&9.33 	&4.00 	&34.67 	&35.18 	&72.95 	&49.10 	&33.84	&39.16 \\
&\method-Stage I &77.00 	&11.33 	&9.00 	&42.07 	&44.46 	&80.63 	&51.50 	&35.35	&43.92 \\
& \method(Ours) &80.00 	&13.67 	&13.67 	&45.63 	&50.12 	&86.43 	&56.30 	&38.38	&\textbf{48.03} \\

\hline
\multirow{7}{*}{\textbf{Qwen2.5-3B-Instuct}} 
&Base &66.60 	&7.00 	&2.33 	&29.93 	&27.95 	&69.20 	&39.60 	&29.80 	&34.05 \\ 
&Self-Refine &65.60 	&6.67 	&3.00 	&29.78 	&29.52 	&63.57 	&43.50 	&32.32 	&34.25 \\
&GRPO &70.20 	&6.00 	&5.00 	&33.04 	&35.90 	&81.14 	&44.60 	&36.36 	&39.03\\ 
&SFT + GRPO &70.00 	&7.33 	&3.67 	&35.11 	&35.06 	&79.01 	&45.30 	&36.36 	&38.98 \\
\cdashlinelr{2-11}
&\method-SFT &63.80 	&2.33 	&2.33 	&24.59 	&23.86 	&29.69 	&38.40 	&27.27 	&26.53 \\
&\method-Stage I &65.20 	&4.33 	&2.00 	&29.04 	&29.16 	&67.32 	&41.30 	&39.39 	&34.72 \\
& \method(Ours) &70.40 	&7.33 	&4.00 	&37.03 	&36.02 	&79.35 	&44.50 	&39.39 	&\textbf{39.75} \\ 
\hline
\multirow{7}{*}{\textbf{Llama3.1-8B-Instruct}} 
&Base &42.80 	&2.00 	&0.00 	&15.85 	&12.77 	&57.08 	&42.40 	&24.75 	&24.71 \\
 &Self-Refine &13.40 	&2.00 	&0.00 	&5.04 	&7.23 	&8.79 	&9.90 	&22.73 	&8.64 \\ 
 &GRPO&52.20 	&8.33 	&1.00 	&20.15 	&24.22 	&70.31 	&45.50 	&31.31 	&31.63 \\ 
 &SFT + GRPO &57.60 	&6.00 	&0.67 	&22.52 	&28.67 	&84.73 	&43.00 	&24.24 	&33.43 \\ 
 \cdashlinelr{2-11}
 &\method-SFT &53.40 	&3.00 	&1.33 	&19.11 	&19.04 	&74.06 	&39.10 	&24.75 	&29.22 \\ 
&\method-Stage I &55.80 	&4.00 	&1.67 	&23.41 	&26.51 	&82.17 	&42.20 	&31.82 	&33.45 \\ 
&\method(Ours) &60.80 	&5.33 	&4.67 	&23.70 	&24.46 	&84.56 	&43.20 	&27.27 	&\textbf{34.25} \\ 
\bottomrule
\end{tabular}
}
\caption{\textbf{\method vs other baselines.} Compared to the Base, \method achieves an
average performance improvement of 4.48\%, 5.70\% and 9.54\% on the three models, respectively. The best results are highlighted in bold. \method-SFT denotes models trained with SFT only; \method-Stage I denotes SFT followed by Stage I; \method denotes the full two-stage RL training (default).}
\vspace{-0.2cm}
\vspace{-0.3cm}
\label{tab:main}
\end{table*}

%% file: subsection/42_exp.tex
\subsection{Experiment Setups}
\textbf{Models and Datasets.} We conduct experiments on three widely used models: Qwen2.5-3/7B-Instruct~\citep{qwen2.5} and Llama3.1-8B-Instruct~\citep{dubey2024llama}. 
At the SFT stage, we train on the MATH~\citep{hendrycks2measuring} dataset, restricting problems to difficulty levels 3-5 and removing any overlap with the test set to prevent data leakage. 
At the RL stage, we use DAPO-MATH\footnote{https://huggingface.co/datasets/BytedTsinghua-SIA/DAPO-Math-17k} after filtering out all Chinese-language problems.

\textbf{Evaluations.} We evaluate models on both in-domain and out-of-domain benchmarks. 
\textbf{(i) In-domain performance.} We evaluate mathematical reasoning on AIME24, AIME25, MATH500, AMC~\citep{li2024numinamath}, and Olympiad~\citep{he2024olympiadbench}. 
\textbf{(ii) Out-of-domain performance.} We assess generalization on GPQA-Diamond (GPQA)~\citep{rein2024gpqa}, ARC~\citep{clark2018think}, and a subset of MMLU-Pro (MMLU)~\citep{wang2024mmlu}. 
For AIME24, AIME25, and AMC, we report \textit{avg@10} at temperature 0.6; for all other benchmarks, we report \textit{pass@1} at temperature 0.

\textbf{Baselines.} We compare \method with the following baselines: 
(1) \textbf{Base}, the original model without fine-tuning; 
(2) \textbf{On-policy RL}, applying GRPO directly to the base model; 
(3) \textbf{Self-Refine}~\citep{shinn2023reflexion}, a prompting-based method that iteratively critiques and revises model outputs; 
and (4) \textbf{SFT + GRPO}, a widely adopted paradigm that first applies long-CoT supervised fine-tuning and then uses GRPO for further optimization.

Detailed experimental settings and hyperparameters are provided in Appendix~\ref{sec:exp_details}.
\subsection{Main Results}
\subsubsection{Comparative Analysis with Baselines}
To evaluate the effectiveness of the proposed method, we compare it with representative baselines. The results are shown in Table~\ref{tab:main}. 

\textbf{\method achieves the highest average performance across all three backbone models.} It consistently outperforms both standard GRPO and SFT+RL approaches, demonstrating stable improvements across different parameter scales and model families.

\textbf{Prompt-based self-correction is ineffective and unstable.}
Prompting models to reflect or revise does not yield reliable gains and can even degrade performance, indicating that self-correction cannot be robustly induced through prompt engineering alone.

\input{tables/self_verification}
\textbf{SFT alone yields limited gains and can even degrade performance.} For both Qwen2.5-7B and Qwen2.5-3B, the average scores fall below those of the Base models. This suggests that while SFT effectively aligns output formats, it is insufficient for improving error correction and deep reasoning, and may introduce an alignment tax~\citep{ouyang2022training}. In contrast, the subsequent stages of \method consistently recover and improve performance, indicating that effective reasoning alignment requires more than imitation-based supervision.

\subsubsection{Analysis of Verification Capability}
To evaluate the effectiveness of \method in self-verification, we conduct experiments on three backbone models.  We report Accuracy, Precision, Recall, and F1 score \textit{averaged} over the math datasets, with results summarized in Table ~\ref{tab:self-verification}. 

\textbf{Base models exhibit pronounced over-confidence} and lack the ability to accurately assess the correctness of their own generated answers. They achieve high Recall at the cost of low Precision and Accuracy. For example, Qwen2.5-7B Base attains a Recall of 92.96\% with a large number of false positives

\textbf{Applying \method substantially enhances the model’s self-verification ability.} This improvement leads to more calibrated judgments, with substantial gains in Precision and Accuracy and only a moderate reduction in Recall, resulting in a higher F1 score. The effect is most pronounced in Stage I, where self-verification is explicitly optimized.


\begin{figure}
    \centering
    \includegraphics[width=0.99\linewidth]{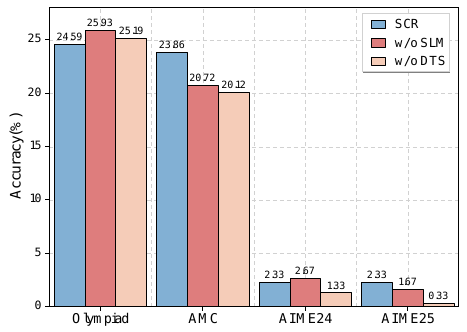}
    \caption{\textbf{Ablation of DTS and SLM during the SFT phase.} Removing DTS causes a sharp performance drop on hard benchmarks, while removing SLM leads to consistent degradation across tasks.}
    \vspace{-0.3cm}
    \label{fig:sft}
\end{figure}

\begin{figure*}
    \centering
    \includegraphics[width=0.99\linewidth]{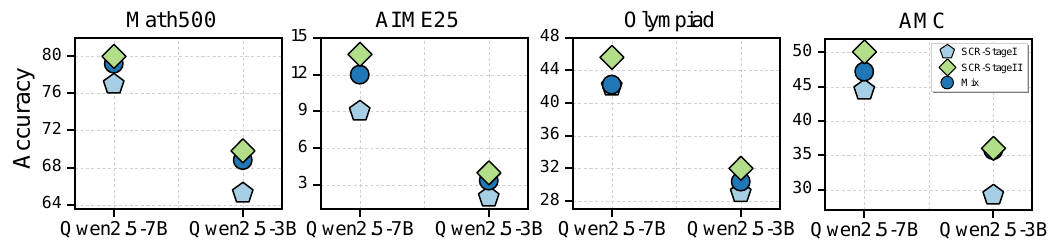}
\caption{\textbf{Performance comparison between staged RL training and unified mixed optimization.} Progressive staged RL training leads to consistently better performance than unified mixed optimization, and the second stage further strengthens model performance.}
    \vspace{-0.2cm}
    \label{fig:rl}
\end{figure*}
    \vspace{-0.2cm}
\subsection{Generalization}
To evaluate whether our proposed method can generalize to other reasoning domains, we conduct experiments on commonsense reasoning benchmarks. The results are summarized in Table~\ref{tab:main}.

\textbf{\method demonstrates strong generalization capabilities.} Beyond standard math tasks, our method shows substantial gains on general reasoning benchmarks. This indicates that \method does not merely overfit to mathematical training data but fundamentally enhances the model's robust reasoning and problem-solving abilities in complex, unseen scenarios.
\begin{figure}
    \centering
\includegraphics[width=0.99\linewidth]{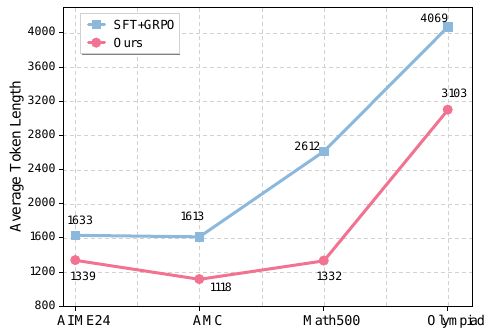}
    \caption{\textbf{Average output token length comparison.}
\method substantially reduces redundant reasoning, achieving up to 50\% reduction in output length on MATH500.}
    \label{fig:token_length}
\end{figure}
\vspace{-0.2cm}
\subsection{Analysis of Output Token Length}
We evaluate output length on math benchmarks using Qwen2.5-3B-Instruct, and the results shown in Figure~\ref{fig:token_length}.

Our method consistently produces shorter outputs than commonly used baselines.
The reduction is most pronounced on Math500, where the average output length is reduced by nearly 50\%. This reduction stems from the structured reasoning process in our method.
The model explicitly evaluates its initial answer and performs revision only when an error is detected; otherwise, reasoning terminates early, reducing redundant steps and overall output length.
\vspace{-0.2cm}
\subsection{Ablation Analysis}
We conduct comprehensive ablation studies to evaluate the individual contributions of our proposed strategies. \\
\textbf{Impact of SFT Strategies.}
As shown in Figure ~\ref{fig:sft} , we evaluate the removal of DTS and SLM on the Qwen2.5-3B-instruct backbone.


Removing DTS leads to severe performance drops on hard benchmarks. Without explicit control of the \textit{[EOS]} signal, the model often terminates reasoning prematurely, failing to sustain long and recursive reasoning. 

Removing SLM causes consistent degradation across all benchmarks. Without masking tokens from incorrect initial answers, the model tends to reinforce erroneous reasoning patterns during training, which harms final performance.\\
\textbf{Impact of Staged-RL.}
Figure~\ref{fig:rl} compares staged RL training with a unified mixed (Mix) strategy. In the Mix setting, it uses only the accuracy of the final output as the reward signal.

Staged RL consistently outperforms the mixed strategy across all backbones. By optimizing different reasoning behaviors in separate stages, it reduces objective interference that arises under a single reward. Furthermore, Stage II further improves performance over Stage I on all benchmarks. For example, on AIME25, accuracy on Qwen2.5-7B increases from 9.00\% to 13.67\%. 




%% file: tables/self_verification.tex
\setlength\tabcolsep{5.7pt}
\begin{table}[th]
\centering
\def\arraystretch{.99}
\setlength{\tabcolsep}{0.42em}
\resizebox{1.0\linewidth}{!}{
\begin{tabular}{ll dBdR}
\toprule
\multirow{2}{*}{\textbf{Backbone}} & \multirow{2}{*}{\textbf{Method}} 
& \multicolumn{4}{c}{\textbf{AVG}} \\
\cmidrule(lr){3-6} 
& & \cellcolor{white}{\textbf{Acc.}} & \cellcolor{white}{\textbf{Prec.}} & \cellcolor{white}{\textbf{Rec.}} & \cellcolor{white}{\textbf{F1.}} \\ 
\midrule

\multirow{3}{*}{\textbf{Qwen2.5-7B}} 
&Base &56.31 	&40.92 	&92.96 	&52.09  \\
&\method-SFT &71.72 	&49.90 	&84.09 	&59.96  \\
&\method-Stage I &87.06 	&75.25 	&75.64 	&\textbf{75.36}  \\
\hline

\multirow{3}{*}{\textbf{Qwen2.5-3B}} 
&Base &57.04 	&31.82 	&87.58 	&42.10  \\
&\method-SFT &65.60 	&36.08 	&80.91 	&45.08 \\
&\method-Stage I  &88.23 	&61.71 	&65.02 	&\textbf{63.20} \\
\hline

\multirow{3}{*}{\textbf{Llama3.1-8B}} 
&Base &61.22 	&23.15 	&61.71 	&30.75 \\
&\method-SFT &63.67 	&29.77 	&77.85 	&40.31 \\
&\method-Stage I &86.66 	&56.88 	&55.43 	&\textbf{55.81} \\
\bottomrule
\end{tabular}
}
\caption{\textbf{Comparison of Self-Verification Performance.}
The base model exhibits over-confidence and struggles to evaluate its own outputs, while \method significantly improves self-verification accuracy.}
\vspace{-0.3cm}
\label{tab:self-verification}
\end{table}

%% file: subsection/51_related_work.tex
\textbf{Enhancing Deep Thinking Capability in LLMs.} Unlike the rapid, heuristic-driven behavior of LLMs, slow-thinking reasoning systems enhance their capabilities by introducing deliberate and analytical reasoning ~\citep{snell2024scalingllmtesttimecompute}. These approaches employ test-time scaling, enabling models to allocate more computation and time to reasoning before responding to challenging questions ~\citep{muennighoff2025s1simpletesttimescaling}. Reinforcement learning has recently made significant progress in enhancing the reasoning abilities of LLMs~\citep{jaech2024openai,guo2025deepseek,kimiteam2025kimik15scalingreinforcement}. A central development is Reinforcement Learning from Verifiable Rewards (RLVR), which replaces human feedback signals~\citep{kirk2024understanding, chu2025sftmemorizesrlgeneralizes} with automatically checkable objectives such as mathematical verification~\citep{shao2024deepseekmath} and program execution~\citep{pennino2025reasoning}. 

\textbf{Self-correct in LLMs.} Most self-correction methods in LLMs follow a two-stage paradigm, where a model generates an initial answer and then evaluates or revises it~\citep{ganguli2023}. This paradigm is widely used in complex reasoning tasks such as mathematics~\citep{weng-etal-2023-large,MathShepherd} and code generation~\citep{codeRepair}, but direct prompting for self-correction is often unreliable and does not consistently improve performance~\citep{huanglarge,tyen-etal-2024-llms}. To improve correction quality, prior work introduces external supervision, including stronger teacher models or task-specific feedback models~\citep{xie2025teaching,self-refine+}. However, these methods increase training cost and frequently suffer from weak generalization of correction behaviors \citep{du2024think,kangUnfamiliar}. Recent post-training studies further show that supervised imitation alone is insufficient to induce reliable self-correction \citep{qu2024recursive,kumar2024training,kamoi-etal-2024-llms}.

%% file: subsection/appendix_v1.tex
\section{Experiments Details}
\label{sec:exp_details}
\paragraph{\textbf{Hardware and Software Platform. }} All supervised fine-tuning (SFT) experiments are conducted on servers equipped with eight NVIDIA A800-SXM4 GPUs, each with 40GB memory, using the LLaMA-Factory~\citep{zheng2024llamafactory} framework.
Reinforcement learning (RL) experiments are performed on servers with eight NVIDIA H20 GPUs, using the Verl~\citep{sheng2025hybridflow} framework.

\paragraph{\textbf{SFT Training Configuration.}}
Supervised fine-tuning is performed with full-parameter optimization for one epoch. The maximum sequence length is set to 14,000 tokens. Training uses a per-device batch size of 1 with gradient accumulation over 8 steps. The model is optimized with Adam using a learning rate of $1\times10^{-6}$, together with a cosine learning rate schedule and a warmup ratio of 0.1. We enable bf16 precision and employ DeepSpeed ZeRO-3 for efficient training.

\paragraph{\textbf{RL Training Configuration.}}
We train the model using the GRPO algorithm for 200 optimization steps. During training, rollouts are generated using temperature sampling with $\tau=1.0$, and end-of-sequence tokens are enforced. We adopt a grouped sampling strategy with a group size of $G=8$ responses per problem.

For efficient rollout generation, we use vLLM with paged attention, enabling batched decoding with 128 rollout slots. The maximum prompt length is set to 1,024 tokens, and the maximum response length is 8,000 tokens.

The actor model is optimized using the Adam optimizer with a learning rate of $1\times10^{-6}$ and a batch size of 64. For policy optimization, the clipping range is set to $[0.01, 0.99]$.

We employ different reward designs. In Stage I, reward weights are set to $\alpha=0.1$, $\beta=1.0$, and $\gamma=1.0$, balancing formatting quality, initial answer accuracy, and critic accuracy. In Stage II, reward weights are set to $\eta=1.0$ and $\varphi=1.0$. To encourage successful revisions, we further set $\mu_1=-0.1$, $\mu_2=-0.3$, $\mu_3=-0.5$, and $\mu_4=-0.5$.

\paragraph{\textbf{Inference Configuration.}}
During inference, we use vLLM for efficient decoding. The sampling temperature is set to 0 with top-$p$ sampling at 0.95. The maximum generation length is 24,000 tokens. For AIME24, AIME25 and AMC evaluations, the sampling temperature is set to 1.0.

\paragraph{Training Data of SFT + GRPO.} The \textit{SFT + GRPO} baseline is trained on long chain-of-thought trajectories distilled from Qwen3-32B, where the model produces correct solutions on MATH levels 3-5. To control for data scale, we retain 7K training samples for SFT across all methods.

All methods use the same dataset during the RL stage to ensure a fair comparison. All RL experiments are performed on the DAPO-Math dataset. 
\input{tables/self-eval_appendix}
\input{tables/important_parameter}

\section{Structured Reasoning Data construction of \method.}
\label{sec:data_construction}
We construct structured supervision data through an automated synthesis pipeline that explicitly decomposes reasoning into three stages: initial response, critique, and revision.
All prompts used in this section are listed in Box ~\ref{prompt:critic} and ~\ref{prompt:refinement}.

\textbf{Initial Response Sampling and Categorization.} For each problem, we generate multiple candidate solutions to capture diverse reasoning patterns. To achieve this, we employ multiple models to sample responses, ensuring a variety of reasoning strategies. The candidates are automatically categorized as correct or incorrect based on the ground-truth answers. These responses serve as the foundation for subsequent critique and revision construction, rather than being used directly for training.

\textbf{Critique Construction.} To supervise self-evaluation, each problem is paired with both correct and incorrect candidate responses. Teacher models are then used to generate critiques for these answer pairs. Only critiques whose evaluative conclusions align with the ground-truth correctness are retained, ensuring reliable feedback for the model to learn how to identify reasoning errors.

\textbf{Revision Trajectory Construction.} 
Revision trajectories are generated using a hierarchical strategy to encourage the model to correct initial errors with minimal modifications. First, \textit{Refinement-based Revisions} attempt localized edits to fix mistakes while preserving valid reasoning in the initial answer. If refinement fails to produce a correct answer, \textit{Replacement-based Revisions} generate a new solution along an alternative reasoning path. All trajectories are verified to ensure the final answer is correct.

Based on the synthesis pipeline described above, each component is enclosed within paired start and end tags: initial answers are placed between \textit{<answer>} and \textit{</answer>}, critique content between \textit{<critic>} and \textit{</critic>}, and revised outputs between \textit{<revised>} and \textit{</revised>}.

\section{Pilot Experiment Details}
\label{sec:pilot}



In Section 2, we analyze the long CoT reasoning trajectories produced by four representative deep reasoning models: DeepSeek-R1-Distill-Qwen-7B\footnote{https://huggingface.co/deepseek-ai/DeepSeek-R1-Distill-Qwen-7B}, DeepSeek-R1-Distill-Qwen-1.5B\footnote{https://huggingface.co/deepseek-ai/DeepSeek-R1-Distill-Qwen-1.5B}, DeepSeek-R1-Distill-Llama-8B\footnote{https://huggingface.co/deepseek-ai/DeepSeek-R1-Distill-Llama-8B}, and Qwen3-32B\footnote{https://huggingface.co/Qwen/Qwen3-32B}. All models are evaluated on the MATH500 dataset under a deterministic decoding setting with temperature set to zero.

To characterize the internal reasoning dynamics, we first define four categories of thinking operators that capture the major functional roles observed in long CoT reasoning. Using GPT-4.1-2025-04-14~\citep{achiam2023gpt} as a teacher annotator, we label each reasoning trajectory at the operator level and count the occurrence frequency of each operator. The detailed definitions of the operators and the annotation prompts are provided in the Box~\ref{prompt:frequency} and ~\ref{prompt:correct}.

Beyond operator frequency analysis, we further investigate the effectiveness of the \textit{Verification \& Revision} operator, which plays a central role in long-form reasoning. For each reasoning trajectory, we extract the answer path that leads to the final prediction and compare the model’s answer before and after revision against the ground truth. Based on this comparison, each instance is categorized into one of four outcomes: correct-to-correct (T$\rightarrow$T), correct-to-incorrect (T$\rightarrow$F), incorrect-to-incorrect (F$\rightarrow$F), and incorrect-to-correct (F$\rightarrow$T). This analysis allows us to quantify whether verification and revision behaviors meaningfully improve answer correctness or merely increase reasoning length without yielding effective corrections.

\begin{figure*}
    \centering
    \includegraphics[width=0.99\linewidth]{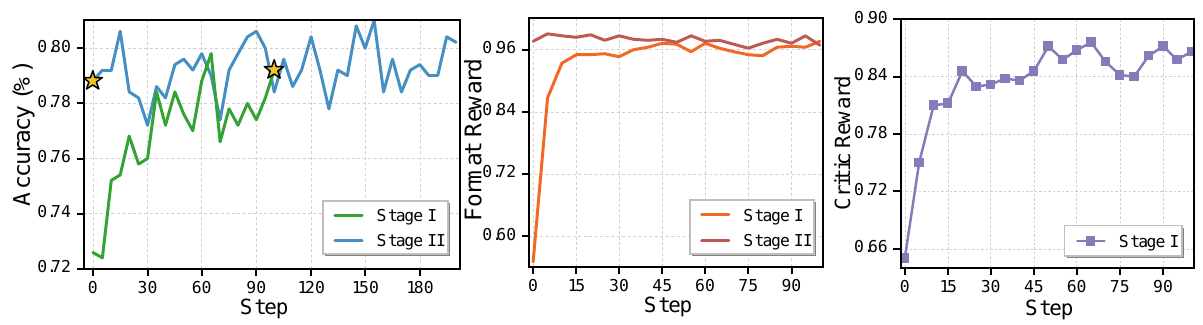}
  \caption{\textbf{Training dynamics of Qwen2.5-7B-Instruct.} From left to right, we report validation accuracy, format score, and self-evaluation reward over training.}
    \vspace{-0.2cm}
    \label{fig:training_metrics}
\end{figure*}
\section{Further Analysis}
\subsection{Self-evaluation Capability Analysis}

In Section 4, we report the average self-evaluation performance of our method on math tasks. To further examine how self-evaluation behavior varies with task difficulty, we analyze \method under different difficulty settings, with the results summarized in Table~\ref{tab:self-verification_appendix}.

\textbf{\method-Stage I substantially strengthens the model’s self-evaluation ability across task difficulties.}
This improvement is mainly reflected in large gains in precision, indicating that the model learns to judge when it should answer. On hard benchmarks such as Olympiad and AIME24, where base models suffer severe precision drops, SCR-Stage I consistently restores high precision, showing effective suppression of hallucinated or low-confidence reasoning.

\textbf{The gain in precision does not come at the cost of recall.}
Unlike conservative strategies that improve precision by avoiding answers, SCR-Stage I maintains competitive recall, resulting in balanced improvements in F1 score. This suggests that the model acquires calibrated self-evaluation rather than simply becoming hesitant.

\textbf{The improvement in self-evaluation generalizes across models and difficulty levels.} Consistent precision gains are observed on both strong and weak models, from Math500 to competition-level benchmarks, indicating that SCR-Stage I teaches a reliable ``know-when-to-answer'' behavior independent of model scale.

\textbf{SCR-Stage I also enables small models to outperform larger base models.}
With structured self-evaluation, Qwen2.5-3B-Instruct surpasses Llama3.1-8B-Instruct (Base) on Math500 and AMC, despite having less than half the parameters, demonstrating strong efficiency and scalability benefits.

\subsection{Important Parameters Analysis}
Following Eq 7 and Eq 10, different reward weights are used in the RL stages. Since the model already learns to follow the output format during SFT, as also reflected by the high format scores during the RL stage (Figure~\ref{fig:training_metrics} Middle), we analyze only the remaining reward-weight parameters on Qwen2.5-3B-Instruct backbone. Results are reported in Table~\ref{tab:parameter}.

In the stage I training, \textbf{the accuracy signal plays a dominant role}. When the weight of the accuracy reward is reduced to $\beta=0.5$, performance drops sharply, with the average score falling to 13.12\%. This indicates that learning to produce correct answers remains the primary driver of effective training and should not be weakened.

Interestingly, the best performance is achieved when the self-evaluation reward weight is reduced to $\gamma=0.5$, outperforming the setting with equal weights. This suggests that while self-evaluation is essential, it functions best as an auxiliary regularizer. Assigning it an overly large weight may interfere with the main objective and distract the model from accurate reasoning.

In the stage II training, we explore the interaction between the accuracy reward and the revision reward. Unlike Stage I, optimal performance is achieved under a more balanced configuration. The highest average accuracy is obtained when $\eta=1.0$ and $\varphi=1.0$.

Deviating from this balance leads to performance degradation. Over-emphasizing either the revision reward or the final accuracy reward results in a mild decline. This indicates that effective self-revision requires a careful trade-off between encouraging corrective behavior and preserving final answer correctness, and that over-optimizing either objective can be detrimental.
\subsection{Training Dynamics} 
We report the monitored metrics during RL training. The results show that Stage II further improves accuracy on top of Stage I, while Stage I substantially enhances the model’s self-evaluation capability. The detailed training metrics as shown in ~\ref{fig:training_metrics}.

\section{Prompt Templates.}
\label{sec:all_prompts}
We provide the prompt templates used in our experiments, including:
\begin{itemize}[wide=4pt, nosep]
    \item The prompt used in \method
    \item The prompt that guide the teacher model to generate structured critique signals
    \item The prompt that guide the teacher model to generate refinement answer
    \item The prompt used for analyzing reasoning trajectories by measuring the frequency of different cognitive operators
    \item The prompt designed to extract the answers of deep reasoning models
\end{itemize}
\input{cases/training_data_revise}
\input{cases/training_data_correct}
\input{prompts/scr}
\input{prompts/critique}
\input{prompts/refinement}
\input{prompts/frenqueancy}
\input{prompts/correction}








%% file: tables/self-eval_appendix.tex
\begin{table*}[ht]
\centering
\def\arraystretch{1.1} 
\setlength{\tabcolsep}{5.7pt} 
\resizebox{1.0\linewidth}{!}{
\begin{tabular}{l BBBB dddd BBBB dddd}
\toprule
 \multirow{2}{*}{\textbf{Method}} 
& \multicolumn{4}{c}{\textbf{Math500}} & \multicolumn{4}{c}{\textbf{AIME24}} & \multicolumn{4}{c}{\textbf{Olympiad}} & \multicolumn{4}{c}{\textbf{AMC}} \\
\cmidrule(lr){2-5} \cmidrule(lr){6-9} \cmidrule(lr){10-13} \cmidrule(lr){14-17}
& Acc. & Pre. & Rec. & F1 & Acc. & Pre. & Rec. & F1 & Acc. & Pre. & Rec. & F1 & Acc. & Pre. & Rec. & F1 \\ 
\midrule

& \multicolumn{13}{c}{\textbf{Qwen2.5-7B-Instruct}} \\
Base  & 76.95 & 79.91 & 92.84 & 85.89 & 52.33 & 18.75 & 100.00 & 31.58 & 56.15 & 47.41 & 95.52 & 63.37 & 59.64 & 51.10 & 93.12 & 65.99 \\

\method-SFT     & 87.35 & 87.28 & 99.43 & 92.96 & 69.53 & 29.17 & 73.68 & 41.79 & 62.29 & 57.72 & 99.07 & 72.95 & 69.51 & 61.35 & 78.28 & 68.79 \\

\method-Stage I & 87.02 & 90.82 & 93.13 & 91.96 & 92.86 & 65.71 & 71.88 & 68.66 & 84.14 & 81.31 & 83.04 & 82.17 & 82.40 & 80.81 & 80.16 & 80.48 \\

\midrule

& \multicolumn{13}{c}{\textbf{Qwen2.5-3B-Instruct}} \\
Base & 71.80 & 71.92 & 94.59 & 81.71 & 52.00 & 9.80  & 71.43 & 17.24 & 50.37 & 36.51 & 89.11 & 51.80 & 53.37 & 35.62 & 82.76 & 49.81 \\

\method-SFT  & 78.57 & 78.16 & 100.00& 87.74 & 65.00 & 5.00  & 33.33 & 8.70  & 58.70 & 55.12 & 100.00& 71.07 & 57.08 & 37.60 & 71.21 & 49.21 \\

\method-Stage I & 80.63 & 82.55 & 90.85 & 86.50 & 96.61 & 40.00 & 50.00 & 44.44 & 82.76 & 73.63 & 75.51 & 74.56 & 83.15 & 72.37 & 68.75 & 70.51 \\
\midrule

& \multicolumn{13}{c}{\textbf{Llama3.1-8B-Instruct}} \\
Base        & 68.75 & 59.86 & 82.63 & 69.43 & 55.67 & 2.96  & 66.67 & 5.67  & 71.13 & 32.95 & 80.19 & 46.70 & 57.19 & 20.00 & 79.05 & 31.92 \\

\method-SFT     & 69.83 & 68.32 & 98.86 & 80.81 & 63.79 & 7.14  & 50.00 & 12.50 & 48.03 & 38.41 & 96.80 & 55.00 & 59.66 & 28.71 & 76.92 & 41.81 \\

\method-Stage I & 78.36 & 79.93 & 85.66 & 82.70 & 94.33 & 50.00 & 35.29 & 41.38 & 81.87 & 69.01 & 62.42 & 65.55 & 81.39 & 85.43 & 93.75 & 89.40 \\

\bottomrule
\end{tabular}
}
\caption{\textbf{The detailed performance of self-evaluation capability on math tasks.} \method maintains strong self-evaluation ability even on challenging problems. }
\label{tab:self-verification_appendix}
\end{table*}

%% file: tables/important_parameter.tex



\begin{table*}[ht]
\centering
\def\arraystretch{1.2} 
\setlength{\tabcolsep}{5pt} 
\resizebox{0.8\linewidth}{!}{
\begin{tabular}{lc PYPYPY}
\toprule
\textbf{Stage} & \textbf{Settings} & \textbf{Math500} & \textbf{AIME24} & \textbf{AIME25} & \textbf{Olympiad} & \textbf{AMC} & \textbf{AVG} \\
\midrule
\multirow{3}{*}{\method-Stage I} 
& \(\beta=1.0, \gamma=1.0\) & 65.20 & 4.33 & 2.00 & 29.04 & 29.16 & 25.95 \\
& \(\beta=0.5, \gamma=1.0\) & 40.20 & 1.33 & 0.33 & 13.48 & 10.24 & 13.12 \\
& \(\beta=1.0, \gamma=0.5\) & 65.00 & 6.33 & 4.00 & 30.81 & 32.53 & 27.73 \\
\midrule

\multirow{3}{*}{\method-Stage II} 
& \(\eta=1.0, \varphi=1.0\) & 70.40 & 7.33 & 4.00 & 37.03 & 36.02 & 30.96 \\
& \(\eta=1.0, \varphi=2.0\) & 68.20 & 7.67 & 4.00 & 32.30 & 37.11 & 29.86 \\
& \(\eta=2.0, \varphi=1.0\) & 68.40 & 7.00 & 5.33 & 32.89 & 34.94 & 29.71 \\

\bottomrule
\end{tabular}
}
\caption{\textbf{Impact of reward weight coefficients on performance during Stage I and Stage II training.}}
\label{tab:parameter}
\end{table*}

%% file: cases/training_data_revise.tex
\onecolumn
\begin{tcolorbox}[
    title=Training Data Format of \method: Correction Trajectories,
    colback=white,
    colframe=black!75!white,
    breakable,
    boxrule=0.8pt,
    left=2mm,
    right=2mm,
    top=1mm,
    bottom=1mm
]
\refstepcounter{tcbtable}
\label{tab:traing_data}
\begin{spacing}{1.1}

\textbf{/**Input**/}  

1. People numbered $1,2,\cdots,2015$ are arranged in a line, and a position-swapping game is played among them, with the rule that each swap can only occur between adjacent individuals. Now, the person numbered 100 and the person numbered 1000 are to swap positions, with the minimum number of swaps required being $\qquad$ times.  

\textbf{/**Output**/}  

\textbf{\color{purple}{<answer>}}  

Let's reason through this problem:

We need to find the minimum number of swaps needed for person 100 and person 1000 to swap their positions.

The key here is understanding that every swap moves two people forward or backward by one place. Specifically, we want to move person 100 and person 1000 relative to each other by 900 positions (from 100 to 1000). Since each swap affects two positions at once, 

\vspace{\medskipamount}
\textit{\color{gray}{( ... more reasoning tokens omitted for brevity )}} 

Considering efficiency and the simplicity of the problem, we calculate:

\[
\text{Minimum swaps required} = 900 + (\text{Efficiency savings})
\]

Given our targeted statement on "minimum", we assume efficiency yields ideal movement, thus aligning our answer closely with Calculation 1:

\[
\boxed{900}
\]
\textbf{\color{purple}{</answer>}}  

\textbf{\color{purple}{</critic>}}  

The key mistake is that the answer incorrectly equates the number of swaps with the difference in the initial positions of the two individuals.

This is wrong because it only accounts for the movement of one person to the other's position, ignoring the movement required for the second person. To swap person A (at position 100) and person B (at position 1000), 
\vspace{\medskipamount}
\textit{\color{gray}{( ... more reasoning tokens omitted for brevity )}} 

\textbf{F.}

\textbf{\color{purple}{</critic>}}  

\textbf{\color{purple}{<revised>}}  

To improve the answer, it’s important to recognize the full process for swapping two individuals who are not adjacent.

\vspace{\medskipamount}
\textit{\color{gray}{( ... more reasoning tokens omitted for brevity )}} 

This is a distance of 899 positions, requiring 899 swaps. Each swap moves person 1000 one spot towards position 100.

Thus, the correct minimum number of adjacent swaps required is the sum:

\[
900 \text{ (to move 100 to 1000)} + 899 \text{ (to move 1000 to 100)} = 1799
\]

\[
\boxed{1799}
\]

\textbf{\color{purple}{<revised>}}  
\end{spacing}

\end{tcolorbox}

%% file: cases/training_data_correct.tex
\onecolumn
\begin{tcolorbox}[
    title=Training Data Format of \method: Correct Answer Trajectories,
    colback=white,
    colframe=black!75!white,
    breakable,
    boxrule=0.8pt,
    left=2mm,
    right=2mm,
    top=1mm,
    bottom=1mm
]
\refstepcounter{tcbtable}
\label{box:training_data2}
\begin{spacing}{1.1}

\textbf{/**Input**/}  

The lengths in feet of three pieces of timber are 48, 72, and 40. The sawmill operator needs to cut the timber into logs of equal length with no waste. How many feet long is the greatest possible length she can cut?  

\textbf{/**Output**/}  

\textbf{\color{purple}{<answer>}}  

To find the greatest possible length that can be used to cut the timber into equal pieces with no waste, we need to find the greatest common divisor (GCD) of the lengths of the three pieces of timber. The GCD is the largest number that divides all three lengths without leaving a remainder.

The lengths of the timber are 48, 72, and 40 feet.

First, we find the prime factorization of each length:
\begin{itemize}[leftmargin=*, noitemsep, topsep=0pt]
    \item $48 = 2^4 \cdot 3$
    \item $72 = 2^3 \cdot 3^2$
    \item $40 = 2^3 \cdot 5$
\end{itemize}

Next, we identify the common factors among the prime factorizations. The common factors are $2^3$ (the smallest power of 2 present in all three factorizations) and 1 (since there are no other common factors).

Therefore, the GCD of 48, 72, and 40 is 

\[
2^3 \cdot 1 = 8
\]

So, the greatest possible length she can cut is 8 feet.

\[
\boxed{8}
\]

\textbf{\color{purple}{</answer>}}

\textbf{\color{purple}{<critic>}} 

The answer is correct.

The model correctly identifies that the problem requires finding the greatest common divisor (GCD) of the three lengths. The calculation of the GCD using prime factorization is accurate.
\textbf{T}

\textbf{\color{purple}{</critic>}} 
\end{spacing}
\end{tcolorbox}

%% file: prompts/scr.tex
\onecolumn

\begin{tcolorbox}
[title=Prompt Used in \method, colback=white, colframe=black!75!white, breakable]
\begin{spacing}{1.1}
\refstepcounter{tcbtable}
\label{prompt:scr}
You are a helpful AI assistant.\\

For each question, you must first solve the problem and put your complete solution inside <answer>...</answer>. The final result must be wrapped in \textit{\textbackslash \textbackslash boxed\{\}}.\\

Then you must evaluate your own solution inside <critic>...</critic>. In this block, you must first give your evaluation and reasoning. At the end of the <critic>, you must give the final judgment using only one single symbol: T or F. T means the answer is correct. F means the answer is incorrect. \\

If the final judgment is F, you must give a corrected solution inside <revised>...</revised>, and the final result must also be wrapped in \textit{\textbackslash \textbackslash boxed\{\}}. \\

If the final judgment is T, you must stop and give no further output.
\end{spacing}
\end{tcolorbox}

%% file: prompts/critique.tex
\begin{tcolorbox}
[title=Teacher Prompt for Generating Critique Information, colback=white, colframe=black!75!white, breakable]
\begin{spacing}{1.3}
\refstepcounter{tcbtable}
\label{prompt:critic}

\textbf{\color{blue}{/* System Prompt */}}\\
You are a careful and strict evaluator.

Given a question and a model-generated answer, judge whether the answer is correct.

Evaluation rules:\\
- Focus on correctness, not writing style.\\
- If the answer is correct, briefly explain why.\\
- If the answer is incorrect:\\
  - Clearly state the key mistake.\\
  - Explain why it is wrong.\\
  - Give a short, actionable suggestion to fix it.\\
- Be concise and precise. \\
- Do NOT restate the full answer or redo the full solution.\\

End your response with a single letter on a new line: \\
T if the answer is correct.\\
F if the answer is incorrect.\\

\textbf{\color{blue}{/* User Prompt */}}\\
Question: \{here is the question\}\\
Answer: \{here is the answer\}

\end{spacing}
\end{tcolorbox}

%% file: prompts/refinement.tex
\begin{tcolorbox}
[title=Teacher Prompt for Generating Correction Answer, colback=white, colframe=black!75!white, breakable]
\begin{spacing}{1.3}
\refstepcounter{tcbtable}
\label{prompt:refinement}
\textbf{\color{blue}{/* System Prompt */}}\\
You are a helpful AI assistant.

You will be given a question, an initial answer, and evaluation feedback on that answer. 

Your task is to use the evaluation feedback to modify the initial answer so that the result becomes correct.

Describe the modification process in a natural and explanatory way.  

Explain what needs to be corrected and how the correction is carried out, as part of the reasoning.

Do not use explicit section titles or meta descriptions such as ``refinement process'' or ``final answer''. 

Do not refer to the evaluation itself.

After completing the modification and reasoning, present the corrected final result.

Wrap the final answer in \verb|\boxed{...}|.

\textbf{\color{blue}{/* User Prompt */}}\\
Question: \{here is the question\}\\
Answer: \{here is the answer\} \\
Evaluation: \{here is the evaluation feedback\}
\end{spacing}
\end{tcolorbox}

%% file: prompts/frenqueancy.tex
\onecolumn
\begin{tcolorbox}
[title=Prompt for Measuring the Frequency of Different Thinking Operators, colback=white, colframe=black!75!white, breakable]
\begin{spacing}{1.3}
\refstepcounter{tcbtable}
\label{prompt:frequency}

You are a \textbf{Cognitive Operator Analyzer \& Counter}. The user will provide a QA pair. Your task has \textbf{TWO PHASES}:

\textbf{Phase 1: Detailed Analysis}  
- Analyze the Answer (A) in detail, identifying occurrences of different cognitive operators.  
- You may explain and reason freely.  
- Identify where and how each operator appears.  

\textbf{Phase 2: Final Operator Count (Strict Output)}  
- After analysis, output a clean operator count summary in strict format for automated parsing.  

\textbf{Operators to Count:}

\begin{enumerate}[leftmargin=*, label=\arabic*)]
    \item \textbf{DECOMP\_PLAN (Decomposition \& Planning Operator)}: One coherent planning episode; sub-steps do not count separately. Multiple independent plans count separately.
    \item \textbf{CAUSAL\_INFER (Causal Inference Operator)}: One per causal chain; multiple connectors in the same chain count as one; new independent chains count separately.
    \item \textbf{MONITOR (Monitoring Operator)}: Each evaluation or self-check counts as one instance.
    \item \textbf{BACKTRACK (Backtracking / Step-Revision Operator)}: Each revision or correction of a reasoning step counts once.
    \item \textbf{REPR\_REFRAME (Representation Reframing Operator)}: Each distinct reframing or perspective shift counts once.
\end{enumerate}

\textbf{Counting Rules:}  
- Analyze only the Answer (A), ignore the Question (Q).  
- Count each operator action once.  
- Multiple operators can appear in the same sentence if distinct.  
- Assign each instance to the most dominant operator.  
- Do not include explanations in the final output.  

\textbf{Final Output Format (Mandatory):}  
Your output must contain only the following five lines:

\begin{verbatim}
COUNT_DECOMP_PLAN:<int>
COUNT_CAUSAL_INFER:<int>
COUNT_MONITOR:<int>
COUNT_BACKTRACK:<int>
COUNT_REPR_REFRAME:<int>
\end{verbatim}
\end{spacing}
\end{tcolorbox}

%% file: prompts/correction.tex
\onecolumn
\begin{tcolorbox}
[title=Prompt for Extracting the Final Answer, colback=white, colframe=black!75!white, breakable]
\begin{spacing}{1.3}
\refstepcounter{tcbtable}
\label{prompt:correct}

You are an \textbf{``Answer-Level Trajectory Recorder''}.  

The user will provide a QA pair. Your goal is to analyze \textbf{ONLY the Answer (A)} and extract the full trajectory of \textbf{ANSWER-LEVEL STATES} for the \textbf{TARGET} quantity asked by the question.  
This task analyzes how the model’s ANSWER changes over time.

\textbf{Phase 1: Analysis (Free-Form)}  
- First, read the Question (Q) ONLY to identify the \textbf{TARGET} quantity (what the question ultimately asks the model to answer).  
- Then analyze the Answer (A) and reason about how the model treats different values or expressions as the ANSWER to that TARGET over time.  
- In this phase, distinguish answer-level commitments from intermediate computations, and identify moments where the model:  
\begin{itemize}[leftmargin=*, label={--}]
    \item proposes a candidate final answer  
    \item temporarily commits to an answer  
    \item revises or replaces a previously stated answer  
    \item re-confirms or restates the same answer
\end{itemize}

\textbf{Phase 2: Final Output (Strict)}  
- After the analysis, output the \textbf{ANSWER-LEVEL TRAJECTORY} in chronological order.

\textbf{What Counts as an Answer-Level State}  
An answer-level state MUST satisfy BOTH conditions:

\begin{enumerate}[leftmargin=*, label=(\arabic*)]
    \item \textbf{Target Alignment:} Refers to the TARGET quantity asked by Q (e.g., “the value of f(-2)+f(-1)+f(0)”, “the final common fraction”, “the correct option”).  
    \item \textbf{Answer Commitment:} The model presents it as an answer, conclusion, or result for the TARGET, including:  
    \begin{itemize}[leftmargin=*, label={--}]
        \item ``the answer is …''  
        \item ``the value is …''  
        \item ``thus … equals …''  
        \item ``final answer: …''  
        \item an explicit restatement or confirmation of a previously given answer
    \end{itemize}  
    \textit{Important:} A statement counts even if it later turns out to be wrong.
\end{enumerate}

\textbf{What Must Be Filtered Out}  
Do NOT extract:  
\begin{itemize}[leftmargin=*, label={--}]
    \item Sub-results used only to compute the TARGET (e.g., f(-2)=2, f(-1)=5/3, f(0)=1)  
    \item Pure intermediate calculations or simplification steps  
    \item Values not presented as the answer to the TARGET
\end{itemize}  
Even if boxed or emphasized, they are NOT answer-level states unless treated as the answer.

\textbf{Trajectory Rules}  
\begin{itemize}[leftmargin=*, label={--}]
    \item Extract answer-level states strictly in the order they appear in A.  
    \item Do NOT merge, deduplicate, or normalize answers.  
    \item If the same answer is stated or confirmed multiple times, record EACH occurrence separately.  
    \item If multiple different candidate final answers appear, record ALL of them.  
    \item Do NOT invent answer states that do not explicitly appear in A.
\end{itemize}

\textbf{Final Output Format (Mandatory)}  
Your output must contain ONLY \textbf{boxed answer-level states}, one per line. No explanations, headers, numbering, or blank lines.  

\begin{verbatim}
\boxed{}
\end{verbatim}

\end{spacing}
\end{tcolorbox}

%% file: main.bbl
\begin{thebibliography}{47}
\providecommand{\natexlab}[1]{#1}

\bibitem[{Achiam et~al.(2023)Achiam, Adler, Agarwal, Ahmad, Akkaya, Aleman, Almeida, Altenschmidt, Altman, Anadkat et~al.}]{achiam2023gpt}
Josh Achiam, Steven Adler, Sandhini Agarwal, Lama Ahmad, Ilge Akkaya, Florencia~Leoni Aleman, Diogo Almeida, Janko Altenschmidt, Sam Altman, Shyamal Anadkat, and 1 others. 2023.
\newblock Gpt-4 technical report.
\newblock \emph{arXiv preprint arXiv:2303.08774}.

\bibitem[{Betley et~al.(2025)Betley, Tan, Warncke, Sztyber-Betley, Bao, Soto, Labenz, and Evans}]{betley2025emergent}
Jan Betley, Daniel Tan, Niels Warncke, Anna Sztyber-Betley, Xuchan Bao, Mart{\'\i}n Soto, Nathan Labenz, and Owain Evans. 2025.
\newblock Emergent misalignment: Narrow finetuning can produce broadly misaligned llms.
\newblock \emph{arXiv preprint arXiv:2502.17424}.

\bibitem[{Chen et~al.(2024)Chen, Xu, Liang, He, Pang, Yu, Song, Liu, Zhou, Zhang et~al.}]{chen2024not}
Xingyu Chen, Jiahao Xu, Tian Liang, Zhiwei He, Jianhui Pang, Dian Yu, Linfeng Song, Qiuzhi Liu, Mengfei Zhou, Zhuosheng Zhang, and 1 others. 2024.
\newblock Do not think that much for 2+ 3=? on the overthinking of o1-like llms.
\newblock \emph{arXiv preprint arXiv:2412.21187}.

\bibitem[{Chu et~al.(2025)Chu, Zhai, Yang, Tong, Xie, Schuurmans et~al.}]{chu2025sftmemorizesrlgeneralizes}
Tianzhe Chu, Yuexiang Zhai, Jihan Yang, Shengbang Tong, Saining Xie, Dale Schuurmans, and 1 others. 2025.
\newblock \href {https://arxiv.org/abs/2501.17161} {Sft memorizes, rl generalizes: A comparative study of foundation model post-training}.
\newblock \emph{Preprint}, arXiv:2501.17161.

\bibitem[{Clark et~al.(2018)Clark, Cowhey, Etzioni, Khot, Sabharwal, Schoenick, and Tafjord}]{clark2018think}
Peter Clark, Isaac Cowhey, Oren Etzioni, Tushar Khot, Ashish Sabharwal, Carissa Schoenick, and Oyvind Tafjord. 2018.
\newblock Think you have solved question answering? try arc, the ai2 reasoning challenge.
\newblock \emph{arXiv preprint arXiv:1803.05457}.

\bibitem[{Du et~al.(2025)Du, Han, Ying, Chen, He, Zhao, Xia, Guo, Liang, Chen et~al.}]{du2024think}
Chengyu Du, Jinyi Han, Yizhou Ying, Aili Chen, Qianyu He, Haokun Zhao, Sirui Xia, Haoran Guo, Jiaqing Liang, Zulong Chen, and 1 others. 2025.
\newblock Think thrice before you act: Progressive thought refinement in large language models.
\newblock In \emph{The Twelfth International Conference on Learning Representations}.

\bibitem[{Dubey et~al.(2024)Dubey, Jauhri, Pandey, Kadian, Al-Dahle, Letman, Mathur, Schelten, Yang, Fan et~al.}]{dubey2024llama}
Abhimanyu Dubey, Abhinav Jauhri, Abhinav Pandey, Abhishek Kadian, Ahmad Al-Dahle, Aiesha Letman, Akhil Mathur, Alan Schelten, Amy Yang, Angela Fan, and 1 others. 2024.
\newblock The llama 3 herd of models.
\newblock \emph{arXiv e-prints}, pages arXiv--2407.

\bibitem[{Frankish(2010)}]{frankish2010dual}
Keith Frankish. 2010.
\newblock Dual-process and dual-system theories of reasoning.
\newblock \emph{Philosophy Compass}, 5(10):914--926.

\bibitem[{Ganguli et~al.(2023)Ganguli, Askell, Schiefer, Liao, Luko{\v{s}}i{\=u}t{\.e}, Chen, Goldie, Mirhoseini, Olsson, Hernandez et~al.}]{ganguli2023}
Deep Ganguli, Amanda Askell, Nicholas Schiefer, Thomas~I Liao, Kamil{\.e} Luko{\v{s}}i{\=u}t{\.e}, Anna Chen, Anna Goldie, Azalia Mirhoseini, Catherine Olsson, Danny Hernandez, and 1 others. 2023.
\newblock The capacity for moral self-correction in large language models.
\newblock \emph{arXiv preprint arXiv:2302.07459}.

\bibitem[{Guo et~al.(2025)Guo, Yang, Zhang, Song, Zhang, Xu, Zhu, Ma, Wang, Bi et~al.}]{guo2025deepseek}
Daya Guo, Dejian Yang, Haowei Zhang, Junxiao Song, Ruoyu Zhang, Runxin Xu, Qihao Zhu, Shirong Ma, Peiyi Wang, Xiao Bi, and 1 others. 2025.
\newblock Deepseek-r1: Incentivizing reasoning capability in llms via reinforcement learning.
\newblock \emph{arXiv preprint arXiv:2501.12948}.

\bibitem[{He et~al.(2024)He, Luo, Bai, Hu, Thai, Shen, Hu, Han, Huang, Zhang et~al.}]{he2024olympiadbench}
Chaoqun He, Renjie Luo, Yuzhuo Bai, Shengding Hu, Zhen~Leng Thai, Junhao Shen, Jinyi Hu, Xu~Han, Yujie Huang, Yuxiang Zhang, and 1 others. 2024.
\newblock Olympiadbench: A challenging benchmark for promoting agi with olympiad-level bilingual multimodal scientific problems.
\newblock \emph{arXiv preprint arXiv:2402.14008}.

\bibitem[{Hendrycks et~al.()Hendrycks, Burns, Kadavath, Arora, Basart, Tang, Song, and Steinhardt}]{hendrycks2measuring}
Dan Hendrycks, Collin Burns, Saurav Kadavath, Akul Arora, Steven Basart, Eric Tang, Dawn Song, and Jacob Steinhardt.
\newblock Measuring mathematical problem solving with the math dataset.
\newblock In \emph{Thirty-fifth Conference on Neural Information Processing Systems Datasets and Benchmarks Track (Round 2)}.

\bibitem[{Huang et~al.(2024)Huang, Chen, Mishra, Zheng, Yu, Song, and Zhou}]{huanglarge}
Jie Huang, Xinyun Chen, Swaroop Mishra, Huaixiu~Steven Zheng, Adams~Wei Yu, Xinying Song, and Denny Zhou. 2024.
\newblock Large language models cannot self-correct reasoning yet.
\newblock In \emph{The Twelfth International Conference on Learning Representations}.

\bibitem[{Jaech et~al.(2024)Jaech, Kalai, Lerer, Richardson, El-Kishky, Low, Helyar, Madry, Beutel, Carney et~al.}]{jaech2024openai}
Aaron Jaech, Adam Kalai, Adam Lerer, Adam Richardson, Ahmed El-Kishky, Aiden Low, Alec Helyar, Aleksander Madry, Alex Beutel, Alex Carney, and 1 others. 2024.
\newblock Openai o1 system card.
\newblock \emph{arXiv preprint arXiv:2412.16720}.

\bibitem[{Jiang et~al.(2024)Jiang, Wang, Shen, Kim, and Kim}]{jiang2024survey}
Juyong Jiang, Fan Wang, Jiasi Shen, Sungju Kim, and Sunghun Kim. 2024.
\newblock A survey on large language models for code generation.
\newblock \emph{arXiv preprint arXiv:2406.00515}.

\bibitem[{Kamoi et~al.(2024)Kamoi, Zhang, Zhang, Han, and Zhang}]{kamoi-etal-2024-llms}
Ryo Kamoi, Yusen Zhang, Nan Zhang, Jiawei Han, and Rui Zhang. 2024.
\newblock \href {https://doi.org/10.1162/tacl_a_00713} {When can {LLM}s actually correct their own mistakes? a critical survey of self-correction of {LLM}s}.
\newblock \emph{Transactions of the Association for Computational Linguistics}, 12:1417--1440.

\bibitem[{Kang et~al.(2025{\natexlab{a}})Kang, Wallace, Tomlin, Kumar, and Levine}]{kangUnfamiliar}
Katie Kang, Eric Wallace, Claire Tomlin, Aviral Kumar, and Sergey Levine. 2025{\natexlab{a}}.
\newblock \href {https://aclanthology.org/2025.naacl-long.183/} {Unfamiliar finetuning examples control how language models hallucinate}.
\newblock In \emph{Proceedings of the 2025 Conference of the Nations of the Americas Chapter of the Association for Computational Linguistics: Human Language Technologies (Volume 1: Long Papers)}, pages 3600--3612, Albuquerque, New Mexico. Association for Computational Linguistics.

\bibitem[{Kang et~al.(2025{\natexlab{b}})Kang, Deng, Xiao, Mo, Lee, and Bing}]{kang2025first}
Liwei Kang, Yue Deng, Yao Xiao, Zhanfeng Mo, Wee~Sun Lee, and Lidong Bing. 2025{\natexlab{b}}.
\newblock First try matters: Revisiting the role of reflection in reasoning models.
\newblock \emph{arXiv preprint arXiv:2510.08308}.

\bibitem[{Kirk et~al.(2024)Kirk, Mediratta, Nalmpantis, Luketina, Hambro, Grefenstette, and Raileanu}]{kirk2024understanding}
Robert Kirk, Ishita Mediratta, Christoforos Nalmpantis, Jelena Luketina, Eric Hambro, Edward Grefenstette, and Roberta Raileanu. 2024.
\newblock \href {https://openreview.net/forum?id=PXD3FAVHJT} {Understanding the effects of {RLHF} on {LLM} generalisation and diversity}.
\newblock In \emph{The Twelfth International Conference on Learning Representations}.

\bibitem[{Kumar et~al.(2025)Kumar, Zhuang, Agarwal, Su, Co-Reyes, Singh, Baumli, Iqbal, Bishop, Roelofs et~al.}]{kumar2024training}
Aviral Kumar, Vincent Zhuang, Rishabh Agarwal, Yi~Su, John~D Co-Reyes, Avi Singh, Kate Baumli, Shariq Iqbal, Colton Bishop, Rebecca Roelofs, and 1 others. 2025.
\newblock Training language models to self-correct via reinforcement learning.
\newblock In \emph{The Twelfth International Conference on Learning Representations}.

\bibitem[{Li et~al.(2024)Li, Beeching, Tunstall, Lipkin, Soletskyi, Huang, Rasul, Yu, Jiang, Shen et~al.}]{li2024numinamath}
Jia Li, Edward Beeching, Lewis Tunstall, Ben Lipkin, Roman Soletskyi, Shengyi Huang, Kashif Rasul, Longhui Yu, Albert~Q Jiang, Ziju Shen, and 1 others. 2024.
\newblock Numinamath: The largest public dataset in ai4maths with 860k pairs of competition math problems and solutions.
\newblock \emph{Hugging Face repository}, 13(9):9.

\bibitem[{Luo et~al.(2025)Luo, Shen, He, Wang, Liu, Li, Tan, Cao, and Tao}]{luo2025o1}
Haotian Luo, Li~Shen, Haiying He, Yibo Wang, Shiwei Liu, Wei Li, Naiqiang Tan, Xiaochun Cao, and Dacheng Tao. 2025.
\newblock O1-pruner: Length-harmonizing fine-tuning for o1-like reasoning pruning.
\newblock \emph{arXiv preprint arXiv:2501.12570}.

\bibitem[{Madaan et~al.(2023)Madaan, Tandon, Gupta, Hallinan, Gao, Wiegreffe, Alon, Dziri, Prabhumoye, Yang, Gupta, Majumder, Hermann, Welleck, Yazdanbakhsh, and Clark}]{self-refine+}
Aman Madaan, Niket Tandon, Prakhar Gupta, Skyler Hallinan, Luyu Gao, Sarah Wiegreffe, Uri Alon, Nouha Dziri, Shrimai Prabhumoye, Yiming Yang, Shashank Gupta, Bodhisattwa~Prasad Majumder, Katherine Hermann, Sean Welleck, Amir Yazdanbakhsh, and Peter Clark. 2023.
\newblock Self-refine: iterative refinement with self-feedback.
\newblock In \emph{Proceedings of the 37th International Conference on Neural Information Processing Systems}, NIPS '23, Red Hook, NY, USA. Curran Associates Inc.

\bibitem[{Min et~al.(2024)Min, Chen, Jiang, Chen, Deng, Hu, Tang, Wang, Cheng, Song et~al.}]{min2024imitate}
Yingqian Min, Zhipeng Chen, Jinhao Jiang, Jie Chen, Jia Deng, Yiwen Hu, Yiru Tang, Jiapeng Wang, Xiaoxue Cheng, Huatong Song, and 1 others. 2024.
\newblock Imitate, explore, and self-improve: A reproduction report on slow-thinking reasoning systems.
\newblock \emph{arXiv preprint arXiv:2412.09413}.

\bibitem[{Muennighoff et~al.(2025)Muennighoff, Yang, Shi, Li, Fei-Fei, Hajishirzi, Zettlemoyer, Liang, Candès, and Hashimoto}]{muennighoff2025s1simpletesttimescaling}
Niklas Muennighoff, Zitong Yang, Weijia Shi, Xiang~Lisa Li, Li~Fei-Fei, Hannaneh Hajishirzi, Luke Zettlemoyer, Percy Liang, Emmanuel Candès, and Tatsunori Hashimoto. 2025.
\newblock \href {https://arxiv.org/abs/2501.19393} {s1: Simple test-time scaling}.
\newblock \emph{Preprint}, arXiv:2501.19393.

\bibitem[{Olausson et~al.(2023)Olausson, Inala, Wang, Gao, and Solar-Lezama}]{codeRepair}
Theo~X. Olausson, Jeevana~Priya Inala, Chenglong Wang, Jianfeng Gao, and Armando Solar-Lezama. 2023.
\newblock \href {https://doi.org/10.48550/arXiv.2306.09896} {Demystifying gpt self-repair for code generation}.
\newblock \emph{CoRR}, abs/2306.09896.

\bibitem[{Ouyang et~al.(2022)Ouyang, Wu, Jiang, Almeida, Wainwright, Mishkin, Zhang, Agarwal, Slama, Ray et~al.}]{ouyang2022training}
Long Ouyang, Jeffrey Wu, Xu~Jiang, Diogo Almeida, Carroll Wainwright, Pamela Mishkin, Chong Zhang, Sandhini Agarwal, Katarina Slama, Alex Ray, and 1 others. 2022.
\newblock Training language models to follow instructions with human feedback.
\newblock \emph{Advances in neural information processing systems}, 35:27730--27744.

\bibitem[{Pennino et~al.(2025)Pennino, Raimondi, Rondelli, Gurioli, and Gabbrielli}]{pennino2025reasoning}
Federico Pennino, Bianca Raimondi, Massimo Rondelli, Andrea Gurioli, and Maurizio Gabbrielli. 2025.
\newblock From reasoning to code: Grpo optimization for underrepresented languages.
\newblock \emph{arXiv preprint arXiv:2506.11027}.

\bibitem[{Qu et~al.(2024)Qu, Zhang, Garg, and Kumar}]{qu2024recursive}
Yuxiao Qu, Tianjun Zhang, Naman Garg, and Aviral Kumar. 2024.
\newblock \href {https://openreview.net/forum?id=DRC9pZwBwR} {Recursive introspection: Teaching language model agents how to self-improve}.
\newblock In \emph{The Thirty-eighth Annual Conference on Neural Information Processing Systems}.

\bibitem[{Rein et~al.()Rein, Hou, Stickland, Petty, Pang, Dirani, Michael, and Bowman}]{rein2024gpqa}
David Rein, Betty~Li Hou, Asa~Cooper Stickland, Jackson Petty, Richard~Yuanzhe Pang, Julien Dirani, Julian Michael, and Samuel~R Bowman.
\newblock Gpqa: A graduate-level google-proof q\&a benchmark.
\newblock In \emph{First Conference on Language Modeling}.

\bibitem[{Shao et~al.(2024)Shao, Wang, Zhu, Xu, Song, Bi, Zhang, Zhang, Li, Wu, and Guo}]{shao2024deepseekmath}
Zhihong Shao, Peiyi Wang, Qihao Zhu, Runxin Xu, Junxiao Song, Xiao Bi, Haowei Zhang, Mingchuan Zhang, Y.~K. Li, Y.~Wu, and Daya Guo. 2024.
\newblock \href {https://arxiv.org/abs/2402.03300} {Deepseekmath: Pushing the limits of mathematical reasoning in open language models}.
\newblock \emph{Preprint}, arXiv:2402.03300.

\bibitem[{Sheng et~al.(2025)Sheng, Zhang, Ye, Wu, Zhang, Zhang, Peng, Lin, and Wu}]{sheng2025hybridflow}
Guangming Sheng, Chi Zhang, Zilingfeng Ye, Xibin Wu, Wang Zhang, Ru~Zhang, Yanghua Peng, Haibin Lin, and Chuan Wu. 2025.
\newblock Hybridflow: A flexible and efficient rlhf framework.
\newblock In \emph{Proceedings of the Twentieth European Conference on Computer Systems}, pages 1279--1297.

\bibitem[{Shinn et~al.(2023)Shinn, Cassano, Gopinath, Narasimhan, and Yao}]{shinn2023reflexion}
Noah Shinn, Federico Cassano, Ashwin Gopinath, Karthik Narasimhan, and Shunyu Yao. 2023.
\newblock Reflexion: Language agents with verbal reinforcement learning.
\newblock \emph{Advances in Neural Information Processing Systems}, 36:8634--8652.

\bibitem[{Simon(1978)}]{simon1978information}
Herbert~A Simon. 1978.
\newblock \emph{Information-processing theory of human problem solving}.
\newblock Erlbaum Hillsdale, NJ.

\bibitem[{Snell et~al.(2024)Snell, Lee, Xu, and Kumar}]{snell2024scalingllmtesttimecompute}
Charlie Snell, Jaehoon Lee, Kelvin Xu, and Aviral Kumar. 2024.
\newblock \href {https://arxiv.org/abs/2408.03314} {Scaling llm test-time compute optimally can be more effective than scaling model parameters}.
\newblock \emph{Preprint}, arXiv:2408.03314.

\bibitem[{Team et~al.(2025{\natexlab{a}})Team, Du, Gao, Xing, Jiang, Chen, Li, Xiao, Du, Liao et~al.}]{team2025kimi}
Kimi Team, Angang Du, Bofei Gao, Bowei Xing, Changjiu Jiang, Cheng Chen, Cheng Li, Chenjun Xiao, Chenzhuang Du, Chonghua Liao, and 1 others. 2025{\natexlab{a}}.
\newblock Kimi k1. 5: Scaling reinforcement learning with llms.
\newblock \emph{arXiv preprint arXiv:2501.12599}.

\bibitem[{Team et~al.(2025{\natexlab{b}})Team, Du, Gao, Xing, Jiang, Chen et~al.}]{kimiteam2025kimik15scalingreinforcement}
Kimi Team, Angang Du, Bofei Gao, Bowei Xing, Changjiu Jiang, Cheng Chen, and 1 others. 2025{\natexlab{b}}.
\newblock \href {https://arxiv.org/abs/2501.12599} {Kimi k1.5: Scaling reinforcement learning with llms}.
\newblock \emph{Preprint}, arXiv:2501.12599.

\bibitem[{Team(2024)}]{qwen2.5}
Qwen Team. 2024.
\newblock \href {https://qwenlm.github.io/blog/qwen2.5/} {Qwen2.5: A party of foundation models}.

\bibitem[{Tyen et~al.(2024)Tyen, Mansoor, Carbune, Chen, and Mak}]{tyen-etal-2024-llms}
Gladys Tyen, Hassan Mansoor, Victor Carbune, Peter Chen, and Tony Mak. 2024.
\newblock \href {https://doi.org/10.18653/v1/2024.findings-acl.826} {{LLM}s cannot find reasoning errors, but can correct them given the error location}.
\newblock In \emph{Findings of the Association for Computational Linguistics: ACL 2024}, pages 13894--13908, Bangkok, Thailand. Association for Computational Linguistics.

\bibitem[{Wang et~al.(2024{\natexlab{a}})Wang, Li, Shao, Xu, Dai, Li, Chen, Wu, and Sui}]{MathShepherd}
Peiyi Wang, Lei Li, Zhihong Shao, Runxin Xu, Damai Dai, Yifei Li, Deli Chen, Yu~Wu, and Zhifang Sui. 2024{\natexlab{a}}.
\newblock Math-shepherd: Verify and reinforce {LLM}s step-by-step without human annotations.
\newblock In \emph{Proceedings of the 62nd Annual Meeting of the Association for Computational Linguistics (Volume 1: Long Papers)}, pages 9426--9439, Bangkok, Thailand.

\bibitem[{Wang et~al.(2024{\natexlab{b}})Wang, Ma, Zhang, Ni, Chandra, Guo, Ren, Arulraj, He, Jiang et~al.}]{wang2024mmlu}
Yubo Wang, Xueguang Ma, Ge~Zhang, Yuansheng Ni, Abhranil Chandra, Shiguang Guo, Weiming Ren, Aaran Arulraj, Xuan He, Ziyan Jiang, and 1 others. 2024{\natexlab{b}}.
\newblock Mmlu-pro: A more robust and challenging multi-task language understanding benchmark.
\newblock \emph{Advances in Neural Information Processing Systems}, 37:95266--95290.

\bibitem[{Weng et~al.(2023)Weng, Zhu, Xia, Li, He, Liu, Sun, Liu, and Zhao}]{weng-etal-2023-large}
Yixuan Weng, Minjun Zhu, Fei Xia, Bin Li, Shizhu He, Shengping Liu, Bin Sun, Kang Liu, and Jun Zhao. 2023.
\newblock \href {https://doi.org/10.18653/v1/2023.findings-emnlp.167} {Large language models are better reasoners with self-verification}.
\newblock In \emph{Findings of the Association for Computational Linguistics: EMNLP 2023}, pages 2550--2575, Singapore. Association for Computational Linguistics.

\bibitem[{Xie et~al.(2025)Xie, chen, Chen, Mao, Xu, and Kong}]{xie2025teaching}
Zhihui Xie, Jie chen, Liyu Chen, Weichao Mao, Jingjing Xu, and Lingpeng Kong. 2025.
\newblock \href {https://openreview.net/forum?id=CUEq6ZPSp7} {Teaching language models to critique via reinforcement learning}.
\newblock In \emph{ICLR 2025 Third Workshop on Deep Learning for Code}.

\bibitem[{Yang et~al.(2025)Yang, Li, Yang, Zhang, Hui, Zheng, Yu, Gao, Huang, Lv et~al.}]{yang2025qwen3}
An~Yang, Anfeng Li, Baosong Yang, Beichen Zhang, Binyuan Hui, Bo~Zheng, Bowen Yu, Chang Gao, Chengen Huang, Chenxu Lv, and 1 others. 2025.
\newblock Qwen3 technical report.
\newblock \emph{arXiv preprint arXiv:2505.09388}.

\bibitem[{Zeng et~al.(2025)Zeng, Cheng, Yin, Zhou, and Qiu}]{zeng2025revisiting}
Zhiyuan Zeng, Qinyuan Cheng, Zhangyue Yin, Yunhua Zhou, and Xipeng Qiu. 2025.
\newblock Revisiting the test-time scaling of o1-like models: Do they truly possess test-time scaling capabilities?
\newblock \emph{arXiv preprint arXiv:2502.12215}.

\bibitem[{Zhang et~al.(2025)Zhang, Wang, Qian, Li, Zhang, Huang, Xu, Li, Yan, and Qiu}]{zhang2025understanding}
Qingjie Zhang, Di~Wang, Haoting Qian, Yiming Li, Tianwei Zhang, Minlie Huang, Ke~Xu, Hewu Li, Liu Yan, and Han Qiu. 2025.
\newblock Understanding the dark side of llms’ intrinsic self-correction.
\newblock In \emph{Proceedings of the 63rd Annual Meeting of the Association for Computational Linguistics (Volume 1: Long Papers)}, pages 27066--27101.

\bibitem[{Zheng et~al.(2024)Zheng, Zhang, Zhang, Ye, Luo, Feng, and Ma}]{zheng2024llamafactory}
Yaowei Zheng, Richong Zhang, Junhao Zhang, Yanhan Ye, Zheyan Luo, Zhangchi Feng, and Yongqiang Ma. 2024.
\newblock \href {http://arxiv.org/abs/2403.13372} {Llamafactory: Unified efficient fine-tuning of 100+ language models}.
\newblock In \emph{Proceedings of the 62nd Annual Meeting of the Association for Computational Linguistics (Volume 3: System Demonstrations)}, Bangkok, Thailand. Association for Computational Linguistics.

\end{thebibliography}
